\documentclass[runningheads]{llncs}
\usepackage{graphicx}
\usepackage{pdfpages}
\usepackage{svg}
\usepackage{colortbl}
\usepackage{adjustbox}
\usepackage{booktabs} 

\usepackage{multirow}
\usepackage{appendix}
\usepackage{xcolor}
\usepackage{bm}
\usepackage{etoolbox}
\usepackage{pifont}
\usepackage{enumitem}
\usepackage{gensymb}
\usepackage{footnote}
\makesavenoteenv{tabular}
\makesavenoteenv{table}

\usepackage{amsmath}
\usepackage{amssymb}
\usepackage{mathtools}

\usepackage[hidelinks]{hyperref}
\usepackage[capitalize]{cleveref}

\usepackage{float}

\newcommand{\ours}{{\tt SPIRL}} 

\newcommand{\f}[2]{\textnormal{#1}(#2)}
\newcommand{\REM}{{\bm E}}
\newcommand{\IND}{\mathbb I}
\newcommand{\nK}{K}
\newcommand{\KP}{O}
\newcommand{\PE}{PE}

\newcommand{\MR}{mr}
\newcommand{\idealMR}{mr^*}

\newcommand{\St}{{\mathcal S}}
\newcommand{\Ac}{{\mathcal A}}
\newcommand{\T}{T}
\newcommand{\R}{R}
\newcommand{\id}{d_0} 
\newcommand{\Expect}{\mathbb E}

\newtoggle{final}
\toggletrue{final}

\newcommand{\TODO}[1]{\iftoggle{final}{}{{\color{red} #1}}}
\newcommand{\pw}[1]{\iftoggle{final}{#1}{{\color{blue} #1}}}
\newcommand{\zj}[1]{\iftoggle{final}{#1}{{\color{orange} #1}}}

\newtoggle{ECMLfinal}
\toggletrue{ECMLfinal}

\newcommand{\ECMLzj}[1]{\iftoggle{ECMLfinal}{#1}{{\color{orange} #1}}}

\newcommand{\citet}[1]{\cite{#1}}
\usepackage[misc]{ifsym}

\begin{document}
\title{Unsupervised Salient Patch Selection for Data-Efficient Reinforcement Learning\thanks{\ECMLzj{Partially supported by the program of National Natural Science Foundation of China (No. 62176154).}}}

%
\titlerunning{SPIRL}
%
\author{\ECMLzj{Zhaohui Jiang\orcidID{0009-0008-5391-5112} \and
Paul Weng\textsuperscript{\Letter}\orcidID{0000-0002-2008-4569}}}
%
\authorrunning{\ECMLzj{Z. Jiang \and P. Weng}}
%
\institute{\ECMLzj{UM-SJTU Joint Institute, Shanghai Jiao Tong University, Shanghai, China\\
\email{\{jiangzhaohui,paul.weng\}@sjtu.edu.cn}}}
\toctitle{Unsupervised Salient Patch Selection for Data-Efficient Reinforcement Learning}
\tocauthor{Zhaohui~Jiang, Paul~Weng}
\maketitle              
\begin{abstract}
To improve the sample efficiency of vision-based deep reinforcement learning (RL), we propose a novel method, called \ours{}, to automatically extract important patches from input images.
Following Masked Auto-Encoders, \ours{} is based on Vision Transformer models pre-trained in a self-supervised fashion to reconstruct images from randomly-sampled patches.
These pre-trained models can then be exploited to detect and select salient patches, defined as hard to reconstruct from neighboring patches.
In RL, the \ours{} agent processes selected salient patches via an attention module. 
We empirically validate \ours{} on Atari games to test its data-efficiency against relevant state-of-the-art methods, including some traditional model-based methods and keypoint-based models.
In addition, we analyze our model's interpretability capabilities.

\end{abstract}
%
%
%
\section{Introduction}\label{sec:intro}

Although deep reinforcement learning (RL) has shown a lot of promise 
\citet{mnih2015DQN,LevineFinnDarrellAbbeel16}, it is notoriously sample inefficient, especially with image inputs.
Indeed, deep RL agents often need to be trained with millions of samples (or even more) before reaching an acceptable performance.
Various techniques (see Related Work in \cref{sec:related}) have been suggested to tackle this issue.
In this paper, we consider two main causes, which we address in our novel method.
First, most deep RL methods learn both end-to-end and from scratch, which is actually not realistic and practical.
Second, the RL agent needs to learn to process a whole image, which may not be necessary since many parts are usually redundant.

Regarding the first point, inspired by recent successes in self-supervised learning in computer vision, we pre-train an adapted Masked Auto-Encoder (MAE) \cite{he2022MAE} to learn a feature extractor via self-supervision.
Recall that in MAE (see Background in \cref{sec:background}), an encoder and decoder, both based on Vision Transformer \cite{dosovitskiy2020ViT}, are jointly trained to reconstruct images from a randomly-sampled subset of its patches.
Our adaptation of MAE help separate the background of an image from its other elements (e.g., moving objects in Atari games) by encouraging the encoder to focus only on those latter elements, while leaving irrelevant parts (e.g., background) to the decoder.
Using such encoder can facilitate the training of the deep RL agent, since the agent does not need to learn to process the raw high-dimensional input anymore and a well-trained encoder should only embed relevant information. 

Regarding the second point, we further exploit the same pre-trained MAE to determine the most salient image patches (i.e., those that are the hardest to reconstruct for the MAE decoder from their surrounding patches).
Interestingly, our salient patch selection method can adaptively determine the number of salient patches depending on the image complexity. 
Our deep RL agent then only takes as inputs the selected salient patches, reducing further the dimensionality of its inputs.
Intuitively, the RL agent may learn faster and better by focusing directly on important parts of an image.

For the deep RL agent itself, we adopt a simple network based on the Transformer architecture \cite{vaswani2017attention} to aggregate the embeddings of the selected salient patches.
This architecture can accept a varying number of salient patches and intuitively learns to exploit their relationships to make decisions.
We call our overall method \emph{Salient Patch Input RL} (\ours{}) and demonstrate the whole approach on Atari games \cite{bellemare2013ALE}.

\paragraph{Contributions}
Our contributions can be summarized as follows: 
%
%
We propose a novel deep RL method (\cref{sec:algo}) that learns to find salient image patches to improve data efficiency\footnote{\ECMLzj{Code is available in \href{https://github.com/AdaptiveAutonomousAgents/SPIRL}{https://github.com/AdaptiveAutonomousAgents/SPIRL}.}}.
Compared to some alternative methods, our model can be pre-trained fast, with a relatively small dataset, and without specifying the number of patches to select.
Moreover, our salient patch selection method is generic and could possibly be used for other downstream tasks.
Since our goal is to improve sample complexity in RL, we propose in addition a simple Transformer-based architecture for the RL agent to process the selected salient patches, whose number can can vary from timestep to timestep.
We experimentally demonstrate (\cref{sec:experiments}) that our method can indeed improve sample efficiency in the low-data regime compared to state-of-the-art (SOTA) methods.
\pw{Finally, we also discuss the interpretability of our method.}

\section{Related Work}\label{sec:related}

Research has been very active in improving the data efficiency of deep RL.
Various generic ideas have been explored, such as 
off-policy training \cite{mnih2015DQN,Lillicrap_Hunt_Pritzel_Heess_Erez_Tassa_Silver_Wierstra_2016,HaarnojaZhouAbbeelLevine18},
improved exploration
\cite{plappert2018parameter,Fortunato_2018},
model-based RL \cite{kaiser2019SimPLe,MBRL},
auxiliary tasks \cite{jaderberg2017reinforcement,Yarats2020}, or
data augmentation \cite{yarats2021image,LaskinLeeStookePintoAbbeelSrinivas20,LinHuangZimmerGuanRojasWeng20} to cite a few.
Most of these propositions are orthogonal to the idea explored in this paper and could be integrated to our method to further improve its performance.
We leave such study to future work.

Closer to our work are techniques proposed to learn more compact representations from images.
We focus mainly on works related to vision-based RL.
They range from keypoint extraction methods to object detection methods.
Our proposed method can be understood as lying at the middle of this spectrum.


Regarding keypoint-based methods, most of them exploit \emph{KeyNet} \cite{jakab2018KeyNet}, which uses a convolutional neural network (CNN) and outputs different keypoints in different CNN channels as Gaussian heatmaps.
For instance, 
\zj{Struct-VRNN \cite{minderer2019StructureVRNN}} or 
\zj{WSDS \cite{ManuelliLiFlorenceTedrake20}}
consider keypoints for learning a dynamic model.
\zj{In addition, \emph{Transporter} \cite{kulkarni2019Transporter} and \emph{PermaKey} \cite{gopalakrishnan2020Permakey}, which both build on KeyNet, improve keypoint detection by considering changes between image pairs and local spatial predictability of CNN features, respectively.}
All these methods need the number $\nK$ of keypoints to be specified before pre-training, which can hinder their performance and applicability in situations where $\nK$ is unknown or dynamic, such as in video games.

Regarding object-based methods, various propositions have been made to learn object-centric representations for RL.
For instance, \citet{agnew2018OLRL} describe a technique notably combining image segmentation and object tracking.
\citet{goel2018MOREL} pre-train a bottleneck-structured network, \ECMLzj{named }\emph{MOREL}, by unsupervised object segmentation with 100k consecutive game frames from a random policy. 
The bottleneck embedding is used as part of the input state for RL policy (PPO or A2C) to reduce the number of interactions with environments.
\citet{zadaianchuk2021SMORL} learn an object-centric latent representation via a compositional generative world model and proves that it can benefit downstream RL manipulation tasks with multiple objects.
These methods are generally more complex, require more hyperparameters to tune, and are slower to train than our approach.




Following the success of 
\ECMLzj{Vision Transformer (ViT) \citet{dosovitskiy2020ViT}} in computer vision, several recent works attempted to exploit it in deep RL.
When applied on convolutional features, \citet{kalantari2022clsQ} report competitive performance in the model-free setting, while \citet{seo2022masked} present a successful model-based approach.
However, current results \cite{Goulao,Tao} seem to suggest that convolutional layers outperform ViT for representation learning in standard RL algorithms, which may not be surprising since Transformer blocks \cite{vaswani2017attention} lack inductive biases like translation invariance encoded in convolutional layers.
To the best of our knowledge, our method is the first demonstrating a promising application of ViT to deep RL without any convolutional layers. 

In the context of vision-based deep RL, prior works used the Transformer model \cite{vaswani2017attention} for the RL agent's decision-making.
For instance, \citet{zambaldi2018RelationalRL} apply such model as a relational inductive bias to process elements of convolutional feature maps.
\citet{kalantari2022clsQ} use such architecture on outputs of a ViT model.
However, there are several differences with our architecture.
Most notably, we use pre-layer normalization instead of post-layer normalization, which has been shown to be inferior \cite{xiong2020PrePostLN,dosovitskiy2020ViT,he2022MAE}.
More importantly, we only process embeddings of salient parts of the whole input image.




\section{Background}\label{sec:background}

We recall next some background on deep RL, the Transformer model, and the masked autoencoder architecture based on Vision Transformer. 

\paragraph{Notations}
Notation $d$ represents an embedding dimension.
For any positive integer $n$, $[n]$ denotes the set $\{1, 2, \ldots, n\}$.
For a sequence $\bm x_1, \bm x_2, \ldots \bm x_n \in \mathbb R^d$, $\bm X = (\bm x_i)_{i \in [n]} \in \mathbb R^{n \times d}$ denotes the matrix stacking them as row vectors.

\subsection{Rainbow and Its Data-Efficient Version}


RL is based on the Markov decision process (MDP) model \cite{Puterman94} defined as a tuple $(\St, \Ac, \T, \R, \id, \gamma)$ where
$\St$ is a state space,
$\Ac$ is an action space,
$\T : \St \times \Ac \times \St \to [0, 1]$ is a transition function,
$\R : \St \times \Ac \to \mathbb R$ is a reward function,
$\id$, is a distribution over initial states, and
$\gamma \in (0, 1)$ is a discount factor.
In this model, the goal is to find a policy $\pi : \St \to \Ac$ such that the expected discounted sum of rewards is maximized.
Formally, this can be achieved by estimating the optimal Q-function defined by:
$
    Q^*(s, a) = \R(s, a) + \gamma \max_{a'} \Expect_{s'}[Q^*(s', a')]
$.

Deep Q-Network (DQN) \cite{mnih2015DQN} approximates this Q-function by minimizing an L2 loss using transitions $(s, a, r, s')$ randomly sampled from a replay buffer storing samples obtained via interactions with the environment:
\begin{equation*}
    (r + \max_{a'} \gamma \hat Q_{\bm\theta'}(s', a') - \hat Q_{\bm\theta}(s, a))^2
\end{equation*}
where $\hat Q_{\bm\theta}$ is the DQN network parametrized by $\bm\theta$ while parameter $\bm\theta'$ is a copy of $\bm\theta$, saved at a lower frequency.

Many techniques have been proposed to improve the efficiency of DQN:
double Q-learning \cite{double},
prioritized replay \cite{prioritized},
dueling network \cite{dueling},
multi-step bootstrap targets \cite{SuttonBarto18},
distributional Q-learning \cite{distributional}, and
noisy DQN \cite{noisy}.
Their combination called \emph{Rainbow} \cite{hessel2018rainbow} is a SOTA value-based method.
By tuning its hyperparameters (i.e., earlier and more frequent updates,  longer multi-step returns), the resulting algorithm called \emph{Data-Efficient (DE) Rainbow} \cite{van2019DERainbow} can be very competitive in the low-data regime, which consists in training with 100K transitions.
In particular, such optimized model-free approach can outperform some model-based methods \cite{van2019DERainbow,kielakrecent2020OTRainbow}.

\subsection{Transformer Layers and Self-Attention}

A \emph{Transformer} model \cite{vaswani2017attention} is defined as a stack of several identical layers, each comprised of a multi-head self-attention (MHSA) module followed by a multi-layer perceptron (MLP) module.
Residual connections \cite{he2016residual} are established around each of them and each module also includes a layer normalization \cite{ba2016LN}.

Formally, one Transformer layer is defined as follows.
For an input sequence of length $n$, $\bm x_1, \ldots, \bm x_n$, its output, $\bm z_1, \ldots, \bm z_n$, can be computed as follows:
\begin{align*}
    \bm x'_i &= \bm x_i + \f{MHSA}{\f{LN}{\bm x_{i}}, \f{LN}{\bm X}} \\
    \bm z_i &= \bm x'_i + \f{MLP}{\f{LN}{\bm x'_{i}}}
\end{align*}
where LN denotes a layer normalization and $\f{LN}{\bm X} = (\f{LN}{\bm x_{j}})_{j \in [n]}$.
Note that we use pre-layer normalization (i.e., LN is applied on the inputs of the two modules, MHSA and MLP), instead of post-layer normalization like in the original Transformer.
The former has been shown to be easier to train and to have better performance \cite{xiong2020PrePostLN,dosovitskiy2020ViT,he2022MAE}.

An MHSA is composed of several scaled dot-product attention (SDPA) heads.
Each SDPA head specifies how an element $\bm x_i$ should attend to other elements of a sequence in $\bm X$.
In practice, several heads are used to allow an element to simultaneously attend to different parts of a sequence.

An SDPA head uses an attention function, which can be understood as a soft associative map.
Formally, given a query $\bm q \in \mathbb R^d$, keys  $\bm K \in \mathbb R^{n \times d}$, and values $\bm V \in \mathbb R^{n \times d}$, this function computes the output associated to $\bm q$ as:
\begin{align*}
    \f{Attention}{\bm q, \bm K, \bm V} = \f{softmax}{\frac{\bm q \bm K^\intercal}{\sqrt{d}}} \bm V \,.
\end{align*}
An SDPA head can then be defined as follows: 
\begin{align*}
     \f{SDPA}{\bm x_i, \bm X} = \f{Attention}{\bm x_i \bm W^Q, \bm X \bm W^K, \bm X \bm W^V} \,,
\end{align*}
where $\bm W^Q, \bm W^K, \bm W^V \in \mathbb R^{d \times d}$ correspond to trainable projection matrices. 
%
An MHSA with $k$ heads is then defined by:
\begin{align*}
    \f{MHSA}{\bm x_i, \bm X} = (\textnormal{SDPA}_\ell(\bm x_i, \bm X))_{\ell \in [k]} \bm W^O \,,
\end{align*}
where the $\textnormal{SDPA}_\ell$'s denote different SDPAs with their own trainable weights, $(\textnormal{SDPA}_\ell(\bm x_i, \bm X))_{\ell \in [k]}$ is seen as a row vector, and $\bm W^O \in \mathbb R^{kd \times d}$ is a trainable projection matrix.

\subsection{Vision Transformer and Masked Autoencoder}

\emph{Vision Transformer} (ViT) \cite{dosovitskiy2020ViT} demonstrates that a Transformer-based architecture without any convolution layer can achieve SOTA performance on computer vision (CV) tasks.
Given an image $\bm X\in\mathbb{R}^{h\times w\times c}$ of height $h$ and width $w$ with $c$ channels, the key idea is to split $\bm X$ into square patches $\{\bm X_{i,j} \in \mathbb{R}^{p^2\times c} \mid (i, j) \in [\frac{w}{p}] \times [\frac{h}{p}] \}$ and linearly project each flattened patches to obtain $\{\bm X^{proj}_{i,j} \in \mathbb{R}^{d} \mid (i, j) \in [\frac{w}{p}] \times [\frac{h}{p}] \}$ as patch embeddings, where $(i,j)$ indicates the patch position in the image, $p$ is the side length of square patches, and $d$ is the embedding dimension for the Transformer layers. 
Since Transformer layers are invariant with respect to permutation of their inputs, positional embeddings are added to patch embeddings before feeding them to the Transformer layers. 

While ViT originally relied on supervised training, 
Masked Auto-Encoder (MAE) \cite{he2022MAE} shows that it can also be trained via self-supervision using image reconstruction from image patches.
MAE is composed of an encoder and decoder, whose architectures are both based on ViT.
Given an image, the encoder receives 25\% randomly selected patches, while the decoder receives the embedding of those patches with a shared trainable [mask] token replacing the 75\% remaining masked patches. 
MAE uses 2-D sinusoidal positional embeddings $\PE_{i,j}$ for position $(i,j)$. 
It therefore feeds $\{\bm X_{i,j}^{proj}+\PE_{i,j} \mid (i, j) \in [\frac{w}{p}] \times [\frac{h}{p}]\}$ to the Transformer layers.

\begin{figure*}[t!]
    \centering
    \includegraphics[trim=0 0 55pt 0, clip,width=.9\textwidth]{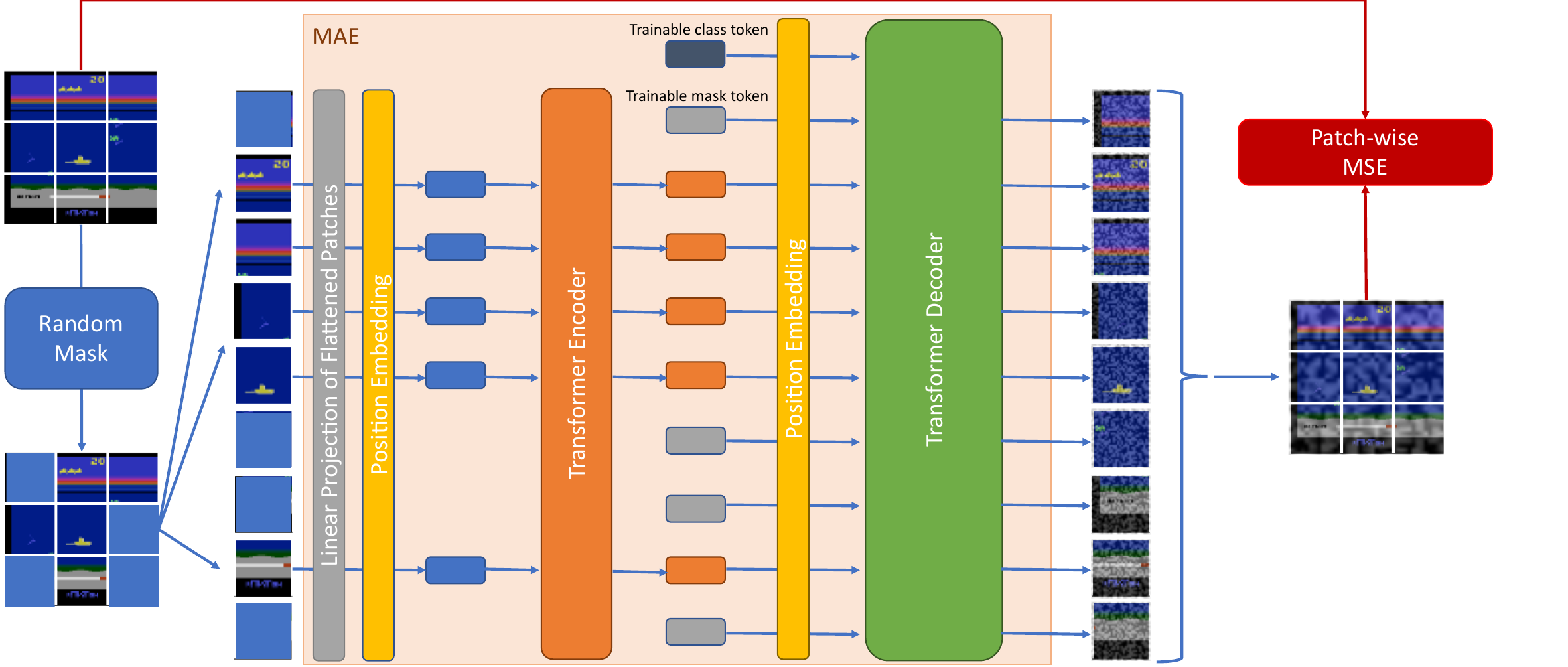}
    \caption{Adaptation of MAE illustrated on \emph{Seaquest}: a higher-capacity decoder can extract the background information, while a smaller encoder focuses on embedding only salient elements in images.
    Overall, its size is more than 50$\times$ smaller than in \cite{he2022MAE}.}
    \label{fig:mae}
\end{figure*}

The MAE model is trained to recover the whole image using a loss defined as a sum over patches of the mean-squared error (MSE) between a normalized patch from the original image and its corresponding patch reconstructed by the decoder.
Formally, for a patch at position $(i, j)$, 
\begin{align}
    &\f{loss}{\bm X_{i,j}, \hat{\bm X}_{i,j}} = \frac{1}{p^2c}\cdot \| \bm X_{i,j}^{norm} - \hat{\bm X}_{i,j}\|^2 \label{equ:MAE_MSE} \\
    &\mbox{where } \bm X_{i,j}^{norm} =\frac{\bm X_{i,j}-\overline{\bm X}_{i,j}}{\hat\sigma_{\bm X_{i,j}}} \label{equ:MAE_per_patch_normalize} \mbox{ and} \\
    &\phantom{where } \hat{\bm X} = \f{MAE}{\f{RandomMask}{\bm X}} \,. \nonumber
\end{align}
Here, $\bm X_{i,j}^{norm}$ is a normalized patch ($\overline{\bm X}_{i,j}$ and $\hat \sigma_{\bm X_{i,j}}$ are the average and standard deviation resp. computed over a patch) and $\hat{\bm X}_{i,j}$ is the corresponding patch extracted from the reconstructed image $\hat{\bm X}$.
%
%
After pre-training, the encoder can be fine-tuned for downstream tasks as a feature extractor.

\section{Architecture}\label{sec:algo}  

\begin{figure}[t!]
    \centering
    \includegraphics[width=.95\columnwidth]{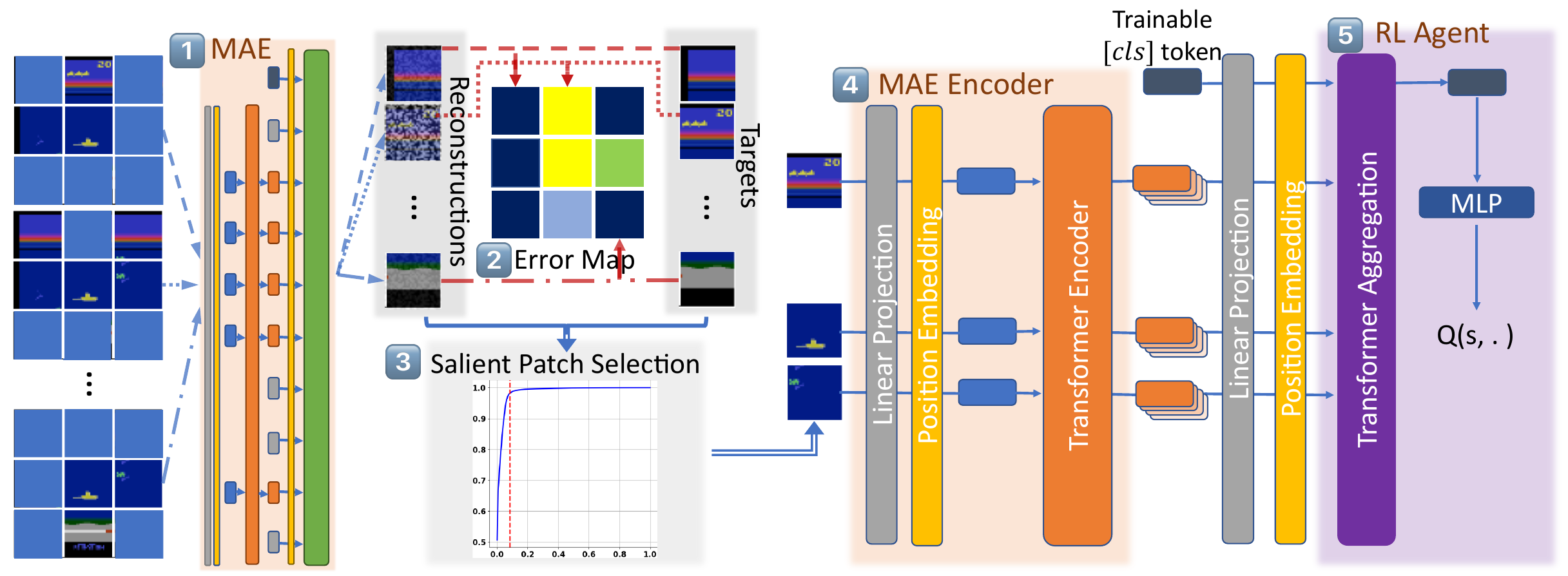}
    \caption{Overview of \ours{}: 
    (1) Pre-trained MAE rebuilds each patch from its surrounding, 
    (2) a reconstruction error map is as MSE between pairs of reconstructed and target patches, 
    (3) most salient patches are selected according to accumulated errors,
    (4) their embeddings are obtained from MAE encoder and concatenated to previous embeddings, and
    (5) A Transformer-based network aggregates them from which an MLP computes the estimated $Q$-values.}
    \label{fig:spirl}
\end{figure}

Our method, called \ours{}, is composed of three main components:
MAE model,
salient patch selection, and
Transformer-based RL model.
See~\zj{\cref{fig:mae,fig:spirl}} for an overview of MAE and \ours{} respectively.
In Atari games, frames are usually square images.
In the remaining, we use $P$ to denote the number of patches per row (i.e., $P=h/p=w/p$).


\subsection{MAE Adaptation}

We show how we specialize the MAE architecture to improve RL data-efficiency and interpretability 
The MAE pre-training is however mostly standard 
(see Appendix A): 
for a given Atari game, we pre-train our MAE network with frames collected from a random policy: 
the MAE encoder learns to encode frame patches and the MAE decoder learns to rebuilt a frame from input patch embeddings.

To facilitate the RL downstream task, we significantly reduce the overall size of our MAE model (see
Appendix A.1
for details) compared to those used in computer vision since Atari frames are smaller and simpler than natural images.
This leads to savings in terms of computational, memory, and sample costs.
Moreover, in contrast to a standard MAE, we define the decoder to be much larger than the encoder for two reasons:
(1) This enforces a stronger bottleneck where the encoder focuses on more essential information intrinsic to a patch, while (2) more global and repetitive information (e.g., background) can be learned by the decoder.
The intuition is that the embeddings returned by the encoder would be easier for the RL agent to learn from.

\begin{figure}[tb!]
    \centering
    \includegraphics[width=.9\textwidth]{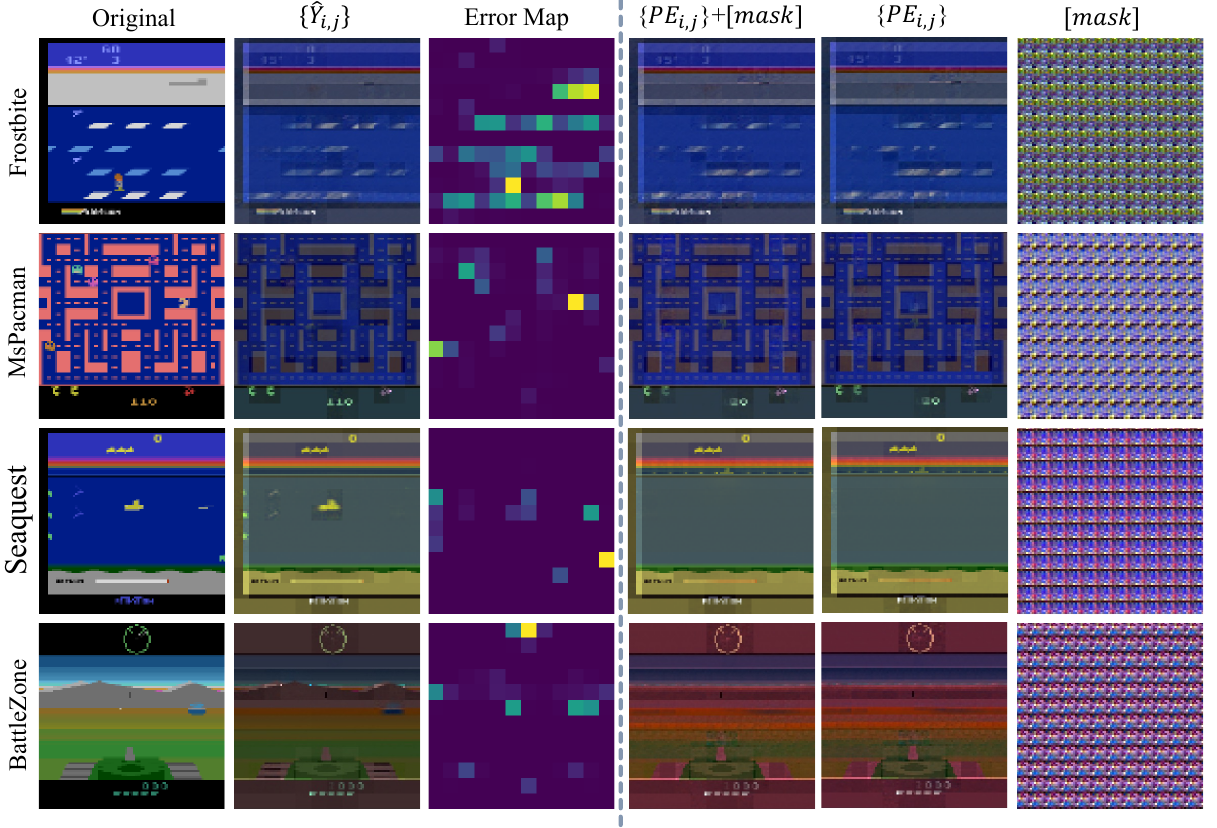}
    \caption{
        \zj{
            Visualization from the pre-trained MAE model.
            The \pw{first} 3 columns are visualizations related to a certain frame:
                (1$^{st}$ column) Original frames;
                (2$^{\pw{nd}}$) Reconstruction from patch surroundings;
                (3$^{\pw{rd}}$) Reconstruction error maps where brighter colors indicate larger values.
            The \pw{last} 3 columns are obtained without frame patches, but with different inputs to the pre-trained MAE decoder: Reconstructions with 
                (4$^{th}$) $\{\PE_{i,j}\}$ added \pw{to} $[mask]$ token;
                (5$^{th}$) only $\{\PE_{i,j}\}$;
                (6$^{th}$) only $[mask]$ token.
        }
    }
    \label{fig:visual_MAE_pretrain}
\end{figure}

We empirically justify those two points on some Atari games.
We start with the second point, which is needed to discuss the first one.
Our architecture can indeed separate the static background from other objects.
\zj{The \pw{last} 3 columns in \cref{fig:visual_MAE_pretrain} shows frames reconstructed by a decoder pre-trained in \pw{a specific game}: \emph{Frostbite}, \emph{MsPacman}, \emph{Seaquest}, and \emph{BattleZone}}
(with frame size $w=h=96$, patch size $p=8$, and thus, $P=12$) for three types of inputs: 
\zj{
(a) positional embeddings alone,
(b) trained $[mask]$ token added with positional embeddings, and
(c) trained $[mask]$ token along.
}
As expected, the decoder can reconstruct the whole background using only the trained \zj{$[mask]$ token} with positional embeddings (i.e., all patches masked, see 
\pw{4$^{th}$ col. of \cref{fig:visual_MAE_pretrain}\pw{)}.}
Interestingly, a finer analysis shows that the positional embeddings alone are necessary 
\pw{
(6$^{th}$ col.)}
and sufficient 
\pw{(5$^{th}$ col.)}
for generating the background, \pw{suggesting} that the background visual information is actually encoded in the parameters of the pre-trained decoder.
%
Note that the colors in the reconstructed frames differ from the original ones \pw{(
1$^{st}$ col.)}, because the MAE model processes normalized inputs (\cref{equ:MAE_per_patch_normalize}).

Regarding the first point, to show that patch embeddings returned by an encoder contain useful local information, we use a pre-trained MAE to reconstruct each patch of a given frame from its surrounding patches as only input.
Formally, for an image patch $\bm X_{i,j}$ at position $(i,j) \in [P]^2$ of an image $\bm X$, we denote its surrounding patches as: 
$
   \f{surr}{\bm X_{i,j}} = \{\bm X_{k,\ell} \!\mid\! (k, l) \in [P]^2, k = i \pm 1, \ell = j \pm 1\} \,.
$
Taking $\f{surr}{\bm X_{i,j}}$ as only input, the pre-trained MAE can reconstruct a whole frame $\f{MAE}{\f{surr}{\bm X_{i,j}}}$, from which we can extract the corresponding reconstructed patch $\hat{\bm  X}_{i,j}$:
\begin{equation} \label{eq:hatX}
    \hat{\bm Y}_{i,j} = \hat{\bm X}_{i, j}\zj{,} \quad\mbox{where } \hat{\bm X} = \f{MAE}{\f{surr}{\bm X_{i,j}}} \, .
\end{equation}
The 
\zj{2$^{nd}$} column of \cref{fig:visual_MAE_pretrain}
shows frames made of the reconstructed $\hat{\bm Y}_{i,j}$ for all $(i, j) \in [P]^2$.
Comparing with the original frames (1$^{st}$ \pw{column}), more accurate details appear at their correct locations in the frames of the \zj{2$^{nd}$ column} than in the background frames generated from the positional embeddings alone (\zj{5$^{th}$ column}).
This confirms that our MAE architecture achieves our first point. 


\subsection{Salient Patch Selection}

As seen in the previous examples, in a given frame, most patches contain redundant or useless information (e.g., containing only static background).
Since an RL agent could potentially learn faster by reducing the dimensionality of its inputs, a natural idea is to discard those irrelevant patches and only keep important ones (i.e., \emph{salient patches}) to formulate the RL agent's input.
We propose to exploit our pre-trained MAE to evaluate the saliency of a patch.
Intuitively, a patch is salient if the decoder cannot reconstruct it from its surrounding patches.
This idea can be formalized by (1) defining a reconstruction error map and (2) using it for selecting salient patches, which we explain next.

\subsubsection{Reconstruction Error Map}

Using a pre-trained MAE, we can calculate a reconstruction error map $\REM\in \mathbb{R}^{P\times P}$ for any image $\bm X$.
Intuitively, the reconstruction error of any patch $\bm X_{i,j}$ of $\bm X$ is obtained by comparing it with its reconstruction $\hat{\bm Y}_{ij}$ (\cref{eq:hatX}) using only its surrounding patches $\f{surr}{\bm X_{ij}}$.
Formally, the reconstruction error map at position $(i,j)$ is simply the MSE between $\bm X_{i,j}$ and $\hat{\bm Y}_{i,j}$:
$
    \REM_{i,j} = \frac{1}{p^2}\|\bm X_{i,j}-\hat{\bm Y}_{i,j}\|^2 \,.
$
Using $\REM$, a patch is deemed more important than another if the $\REM$ score of the former is larger than 
latter.

As an illustration, \zj{the 2$^{nd}$ column in \cref{fig:visual_MAE_pretrain}} shows frames made of the reconstructed $\hat{\bm Y}_{i,j}$ for all $(i,j)\in[P]^2$, and
their corresponding reconstruction error maps (\zj{3$^{rd}$ column}) where dark blue corresponds to smaller errors, while green (resp. yellow) corresponds to larger (resp. largest) errors.
Interestingly, patches with larger errors roughly correspond to moving objects or objects with a complex hard-to-predict surface (i.e., 
Pacman, ghosts, floating ices, fishes, submarine, game score, or life count associated with higher error values), suggesting that the reconstruction error map can be exploited for salient patch selection.

\begin{figure}[t!]
    \centering
    \begin{minipage}[t]{.48\textwidth}
        \centering
        \includegraphics[width=1\linewidth, height=0.42\textheight]{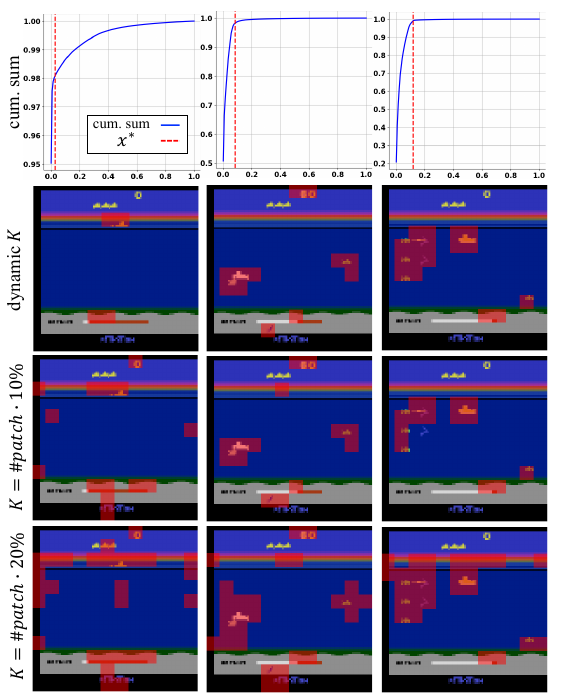}
        \caption{Different salient patch selection strategies in three different frames \pw{(one per column)} of \textit{Seaquest}: 
            (1$^{st}$ row) 
            \pw{blue lines (resp. red dashed lines) are cumulative sum of errors in $\REM$ (resp. $p^*$);}
            (2$^{nd}$) selected patches (red squares) with $\nK$ determined by $p^*$ from  1$^{st}$ row; 
            (3$^{rd}$ \& 4$^{th}$) selected patches \pw{with} pre-defined $\nK$. 
            }
        \label{fig:cumsum_45_mr}
    \end{minipage}%
    \hspace{0.3cm}
    \begin{minipage}[t]{0.48\textwidth}
        \centering
        \includegraphics[width=1\linewidth, height=0.42\textheight]{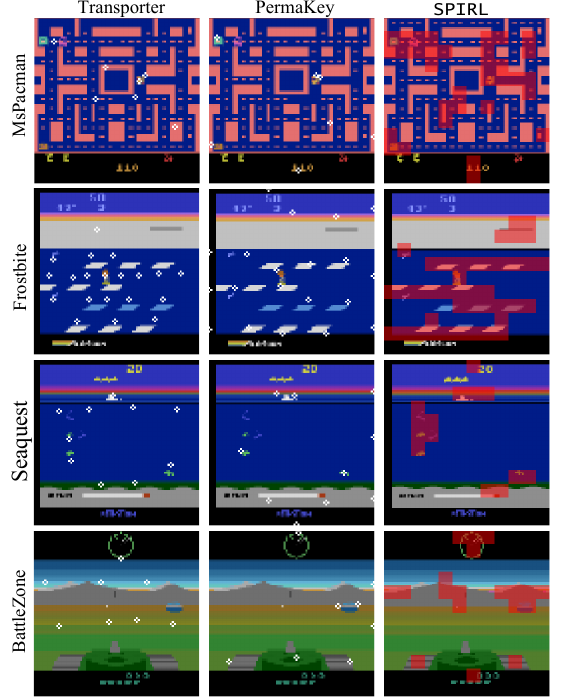}
        \caption{Comparison of key-point (represented as white crosses) / salient-patch (transparent red patches) selection using Transporter, PermaKey, or \ours{}. \zj{Each row corresponds to a same frame \pw{of a given game}. 
        Visualization for \pw{other} frames can be \pw{found} in 
        Appendix A.4.}}
        \label{fig:compare_kpts}
    \end{minipage}
\end{figure}

\subsubsection{Salient Patch Selection}

Using the reconstruction error map $\REM$, salient patches can be selected, but one still needs to decide how many to keep.
A simple approach, as done in all previous works \cite{jakab2018KeyNet,minderer2019StructureVRNN,kulkarni2019Transporter,gopalakrishnan2020Permakey}, would be to set a fixed pre-defined $\nK \in \mathbb N$ and select the top $\nK$ in terms of $\REM$ scores.
However, a suitable value for $\nK$ is not easy to choose: if too low, important information will be missing, but if too high, too many patches would need to be processed by the RL agent.
Moreover, a pre-defined $\nK$ may be too simplistic, because the number of salient patches usually vary from games to games, but also from frames to frames for a given game.
For instance, as \cref{fig:cumsum_45_mr} shows in \emph{Seaquest}, \pw{objects may appear or disappear}. 
Each column in \cref{fig:cumsum_45_mr} represents a frame at a different timestep and the 3$^{rd}$ and 4$^{th}$ rows depict the selected key frames for two different $\nK$'s expressed as a percentage of the total number of patches.

To be more adaptive, we propose instead a selection method, called \emph{dynamic $\nK$}.
A natural way to express it is via the absolute Lorenz curve \cite{KOWejor04} of $\REM$.
\pw{To define it formally, we first recall the cumulative distribution of the errors:}
$
    F(x) = \frac{1}{P^2}\sum_{i,j} \IND_{\REM_{i,j} \le x} \,,
$
where $\IND$ is the indicator function.
Consider its left-con\-ti\-nu\-ous inverse (i.e., quantile function) for $p \in (0, 1]$:
$
    F^{-1}(p) = \inf \{x \in [0, 1] \mid F(x) \ge p\} \,.
$
The absolute Lorenz curve $F^{-2}$ is defined as its integral:
$
    F^{-2}(x) = \int_0^x F^{-1}(p) dp \,.
$
This curve, known to be concave, describes the allocation of errors over the population of patches.
If this allocation were uniform, the curve would be a straight line from point $(0, 0)$ to $(1, 1)$.
As expected, in practice, it is far from uniform (1$^{st}$ row of \cref{fig:cumsum_45_mr}).
There is a diminishing return for selecting more salient patches.
We express the problem of selecting the best $\nK$ as finding $p$ such that the increase in the accumulated errors start to decrease.
Formally, this point is approximately the point $(p^*, x^*)$ such that its slope is closest to $45$°.

   
   
    



\subsection{Transformed-based RL}

After the dynamic salient patch selection, $\nK_t$ patches whose embeddings are given by the pre-trained embeddings are used as input for the downstream RL tasks.
Note that we now index $\nK$ with time since this number can vary from frame to frame.
We denote the set of selected salient patches at time $t$ as $\KP_t$.

The problem of how the embeddings in $\KP_t$ should be processed by the RL agent is not completely straightforward. 
Indeed, since our method does not guarantee temporal consistency across different frames (e.g., number or order of selected salient patches), the simple approach consisting in concatenating the embeddings is not feasible, in contrast to previous methods such as
\zj{Transporter \cite{kulkarni2019Transporter} or PermaKey \cite{gopalakrishnan2020Permakey}}.
In addition, concatenation is also not practical when $\nK$ becomes large.
Moreover, a direct pooling (e.g., via a mean or $\max$) over these embeddings would be a too lossy operation.


To solve this problem and obtain an architecture that is invariant with respect to permutations of its inputs, we propose the following simple Transformer-based architecture to aggregate the embeddings.
A linear trainable projection is applied to each embeddings of $\KP_t$ to reduce their dimension.
The resulting embeddings are then processed by several Transformer layers to obtain a global final embedding $\bm o_t$, defined as $\f{MHSA}{[cls], \bm X_t}$ where $[cls]$ is a trainable class token and $\bm X_t$ contains the embeddings of patches in $O_t$.
Embedding $\bm o_t$ serves as input for the final decision-making part.

Although our method based on MAE and salient patch selection is generic, we instantiate it with DE Rainbow for simplicity.
Therefore, following previous practice, the RL agent processes a concatenation of the four latest $\bm o_t$ to account for partial observability using a simple MLP for the final decision-making part.
Our overall proposition can be understood as replacing the convolutional layers of a standard Rainbow network by three components: 
a pre-trained MAE encoder, 
the dynamic $\nK$ salient patch selection, and
MHSA layers.

\begin{table*}[t!]
\caption{(Left) Pre-training time cost of \ours{} and baselines (averaged over the 4 Atari games); (Right) 
\pw{\ours{} with same configurations as in \cref{tab:exp_results_atari}-\emph{100K} v.s. MAE-All.}
}
\label{tab:ptrtraining_time_cost}
\label{tab:abla_vitAll}
\centering
    \begin{minipage}[t]{0.25\textwidth}
    \centering
    \begin{small}
    \begin{adjustbox}{max width=\textwidth}
        \begin{tabular}{lc}
        \toprule
        Method   & Time (hours) \\
        \midrule
        Transporter & 25.9 \\
        PermaKey & 50.7 \\ 
        \ours{} & 1.6\\
        \bottomrule
        \end{tabular}
    \end{adjustbox}
    \end{small}
    \end{minipage}%
    \hspace{0.1cm}
    \begin{minipage}[t]{0.73\textwidth}
    \centering
    \begin{small}
    \begin{adjustbox}{max width=\textwidth}
        \begin{tabular}{lcccc}
        \toprule
        Inputs for RL & Frostbite    & MsPacman     & Seaquest     & BattleZone      \\
        \midrule
        \ours{}         & \textbf{669.5(731.7)} & \textbf{904.1(286.4)} & \textbf{557.9(148.1)} & \textbf{11980.0(3826.7)} \\
        MAE-All       & 405.7(382.3) & 760.3(210.0) & 538.4(79.8)  & 10628.0(3176.9)\\
        \bottomrule
        \end{tabular}
    \end{adjustbox}
    \end{small}
\end{minipage}
\ECMLzj{\vspace{-0.3cm}}
\end{table*}

\section{Experimental Results}\label{sec:experiments}
For the pre-training part, we mainly compare \ours{} against counterparts including Transporter \cite{kulkarni2019Transporter}, PermaKey \cite{gopalakrishnan2020Permakey}, and MOREL \cite{goel2018MOREL}, which also use pre-training to learn object-centric features to improve RL performance. 
For the RL training part, we focus on comparing with SOTA methods evaluated in the low-data regime, including SimPLe \cite{kaiser2019SimPLe}, Transporter, PermaKey. \zj{We further show examples about using attention to explain the learned policy with \ours{}.}


\subsubsection{Pre-training}

\ours{} costs less than 50K interaction steps to collect pre-training data from a random policy, while other baselines either need to collect frames with trained agents or need at least 1.5 times more training data \zj{with higher or equal data quality requirement} than us (
Appendix A.2
). 
Although \ours{} is less demanding in terms of data quality and quantity for pre-training, its salient patch selection is qualitatively at least as good as other baselines (\cref{fig:compare_kpts}). 
While counterparts may not assign keypoints to important objects successfully (e.g., ghosts in \textit{MsPacman}, small and semitransparent fishes in \textit{Seaquest}, tiny bullet in the center in \textit{BattleZone}), \ours{} manages to
capture such patches correctly.

\cref{tab:ptrtraining_time_cost} (Left) shows that \ours{} can be pre-trained much faster than Transporter and PermaKey when using a same machine configuration (Appendix A.3). 

Once our pre-training is finished, the reconstruction error map can be used to select varying numbers of salient patches in downstream tasks, which amortize the pre-training cost further.
Recall that in contrast, previous keypoint methods would need to pre-train again if $\nK$ is changed.


\begin{table*}[t!]
\caption{\ours{} v.s. baselines on Atari environments (Appendix B.4
for result sources).  
}
\label{tab:exp_results_atari}
\centering
\begin{small}
\begin{adjustbox}{max width=\textwidth}
    \begin{tabular}{lcccccccccc}
    \toprule
    \multirow{2}{*}{Games}  & \multicolumn{5}{c}{\emph{100K}, average} & \multicolumn{2}{c}{\emph{400K}, median}\\ \cmidrule(lr){2-6} \cmidrule(lr){7-8}
    & SimPLe & Transporter & DE-Rainbow & \zj{DE-Rainbow-P} & \ours{} & PermaKey & \ours{}\\ 
    \midrule
    Frostbite  & 254.7 & 388.3(142.1) & 341.4(277.8) & 483.7(710.7) & \textbf{669.5(731.7)} & 657.3(556.8) & \textbf{1194.0(1283.8)}\\
    MsPacman   & 762.8 & 999.4(145.4) & \textbf{1015.2(124.3)} & 985.4(129.6) & 904.1(286.4) & 1038.5(417.1) & \textbf{1186.0(193.9)}\\
    Seaquest   & 370.9 & 236.7(22.2)  & 395.6(124.4) & 462.5(187.8) &  \textbf{557.9(148.1)} & 520.0(169.9) & \textbf{534.0(88.2)}\\
    BattleZone & 5184.4 & —— & 10602.2(2299.6) & \textbf{13992.0(4713.5)} & 11980.0(3826.7) & 12566.7(3297.9) & \textbf{13500.0(1870.8)}\\
    \bottomrule
    \end{tabular}
\end{adjustbox}
\end{small}
\end{table*}



\subsubsection{Atari Benchmark}

We test \ours{} on Atari games \cite{bellemare2013ALE}, \pw{a commonly-used RL benchmark}, to demonstrate the performance of our framework. 
We choose the four games, \textit{MsPacman}, \textit{Frostbite}, \textit{Seaquest}, and \textit{BattleZone}, because they have various environment properties to demonstrate the ability of a data-efficient algorithm and they have also been adopted in PermaKey.
We keep the hyper-parameters used in DE-Rainbow \cite{van2019DERainbow} (see 
Appendix B.2). 

To illustrate \ours{}'s data-efficiency, we focus on training in the low-data regime, which was already used to evaluate SimPLe.
In this training regime, denoted \emph{100K}, agents are trained with only 100K environment steps (i.e., 400K frames, since 4 frames are skipped every step). 
\pw{We also evaluate DE-Rainbow (denoted DE-Rainbow-P) trained with replay buffer initially filled with the same trajectory data as we pre-train our MAE before learning (see
Appendix B.4
for more configuration details).}
Note that Permakey uses a variant of this low-data training regime, denoted \emph{400K}: 
it allows the agent to interact with the environment 400K steps, but only updates the network 100K times.
Although both regimes use the same number of updates, the latter one is more favorable and allows the agent to generate more data from more recent and better policies.

\zj{
Since the number of selected salient patches may vary \pw{in an episode}, 
we can not train the RL agent by sampling a batch of data, since the length of data inside a batch is not aligned. Moreover, diverging number of patches make\pw{s} the training unstable and hurt\pw{s} the performance.
As an implementation trick, in order to stabilize training and reduce the size of the replay buffer, we determine a maximal ratio $\MR$ for the number of the selected patches based on counts of numbers of salient patches from the pre-training datasets. 
If more than $\MR$ salient patches are selected, we simply drop the extra patches with the smallest error values. Otherwise we add dummy patches with zero padding.
Detailed architecture and method to select $\MR$ can be checked in Appendix B.5.
}%

\cref{tab:exp_results_atari} compares \ours{} against other baselines. 
To make the results comparable, we run \ours{} according to the evaluation settings used in the baselines: 
average score for \emph{100K} evaluated at the end of training, and the best median score for \emph{400K} evaluated with the best policy during training. 
Further evaluation details are given in Appendix B.3.
\ours{} can outperform Transporter and PermaKey \zj{without \pw{the need of tuning} the number of selected patches in a wide range of candidates for different games.}
For \textit{MsPacman} and \textit{BattleZone}, \ours{}'s performance is not as good as DE-Rainbow but is still reasonable. 
For \textit{MsPacman}, 
the lower performance can be explained by the importance of the background, which we remove in \ours{} by design.
Since we pre-train in MsPacman with a random policy, pebbles and corridors in the maze will be usually regarded as background. 
The performance of \ours{}, but also all key-point methods, will be limited if the background is important for decision-making.
For \textit{BattleZone}, 
\zj{DE-Rainbow-P can benefit a lot from richer transition data from replay buffer, while the performance decrease of \ours{}}
comes from the moving mountain backgrounds, which makes \ours{} select more irrelevant background patches than needed.
For this kind of situation, the performance of \ours{} can be improved by using an optimized fixed pre-defined $\nK$ or tune the dynamic $\nK$ criterion (e.g., angle smaller than $45$°).




\subsubsection{Ablation Study} \label{sec:ablation}
\zj{
\pw{To understand the effect of our patch selection, we compare \ours{} with MAE-All where the RL agent processes all patch embeddings.}
\cref{tab:abla_vitAll} (Right) shows that selecting salient patches improve\pw{s} \pw{RL} data-efficiency in low-data regime. 
Visualization of policy attention \pw{(\cref{fig:interpretability_main})} indicates that \pw{in contrast to \ours{}, }MAE-All \pw{attends non-necessary patches, hurting its} performance.}


\begin{table}[tb!]
\caption{\zj{Ablation study to compare dynamic-$\nK$ (abbr. D-$\nK$) with $\MR$ or simply using determined-$\nK$ on \emph{100K} with average scores.  
\pw{(best results for each game in bold).}
}}
\label{tab:compare_mmr_mr}
    \begin{adjustbox}{max width=\textwidth}
        \begin{tabular}{lcccccc}
        \toprule
        Games                 & \multicolumn{3}{c}{Frsotbite}                                & \multicolumn{3}{c}{MsPacman}                                      \\
        \cmidrule(lr){2-4} \cmidrule(lr){5-7}
        ratio x                         & $\idealMR$+5\%       & $\idealMR$=60\%               & $\idealMR$-5\%      & $\idealMR$+5\%         & x*=65\%                  & $\idealMR$-5\%      \\
        \midrule
        D-$\nK$, $\MR$=x & 266.8(109.8) & 350.6(350.9) & \textbf{669.5(731.7)}  & 847.9(314.3)   & 735.9(196.8)             & \textbf{904.1(286.4)} \\
        $\nK$ = $P^2\times x$ & 410.6(373.6) & 565.4(538.7)          & 563.0(720.7)   & 859.3(311.2)   & 909.2(512.2)             & 710.5(146.4)  \\
        \midrule
        Games               &
        \multicolumn{3}{c}{Seaquest}                                 & \multicolumn{3}{c}{BattleZone}                                    \\
        \cmidrule(lr){2-4} \cmidrule(lr){5-7}
        ratio x                         & $\idealMR$+5\%       & $\idealMR$=80\%               & $\idealMR$-5\%    & $\idealMR$+5\%         & $\idealMR$=70\%                  & $\idealMR$-5\%        \\
        \midrule
        D-$\nK$, $\MR$=x & 371.6(89.5)  & \textbf{557.9(148.1)} & 464.8(78.3)   & 8624.0(2667.1) & \textbf{11980.0(3826.7)} & 9660.0(3462.8)   \\
        $\nK$ = $P^2\times x$ & 478.0(133.6) & 496.3(90.1)           & 438.1(156.0)   & 8035.3(2173.3) & 8700.0(2173.3)           & 7600.0(3580.6)  \\
        \bottomrule
        \end{tabular}
    \end{adjustbox}
\end{table}

\zj{
\pw{To understand Dynamic-$\nK$ (implemented with the $\MR$ trick using zero-padding for unneeded patches), we compare \ours{} with fixed $\nK$ determined by $\MR$, i.e., $\nK=\#patches\times\MR$ patch embeddings provided} to RL policy. 
\cref{tab:compare_mmr_mr} \pw{shows} that our dynamic-$\nK$ method performs better.
}

We \zj{also} tested some variations in the \ours{} architecture and implementation details.
\pw{See Appendix C.1 and C.2
for those experimental results.}


\zj{
\subsubsection{Policy Interpretability \pw{via} Policy Attention}
The learned policy can be interpreted \pw{to} some \pw{extent} from the policy attention score\pw{s} \pw{w.r.t.} $[cls]$ token, since we use $[cls]$ to pool the global information for downstream RL part. \cref{fig:interpretability_main} visualize\pw{s} the policy-attended patches by maintaining $60\%$ mass of $\f{softmax}{\frac{\bm q^{[cls]} \bm K^\intercal}{\sqrt{d}}}$.
As illustrated in the top 2 rows in \cref{fig:interpretability_main}, \pw{the} trained policy attend\pw{s} patches with useful objects (e.g.\pw{,} submarine, oxygen bar, \pw{enemies}) than non-necessary patches (e.g.\pw{,} patches with the top rainbow bar), 
\pw{which explains why the shooting action is chosen (selected patches with enemy on the right).}
Moreover, by checking the policy-attention from a trained MAE-All policy as shown in the 3$^{rd}$ row in \cref{fig:interpretability_main}, we find non-necessary policy-attention are assigned to non-salient patches, which could explain why MAE-All perform worse than \ours{}.  

\begin{figure}[tb!]
    \centering
    \includegraphics[width=\textwidth]{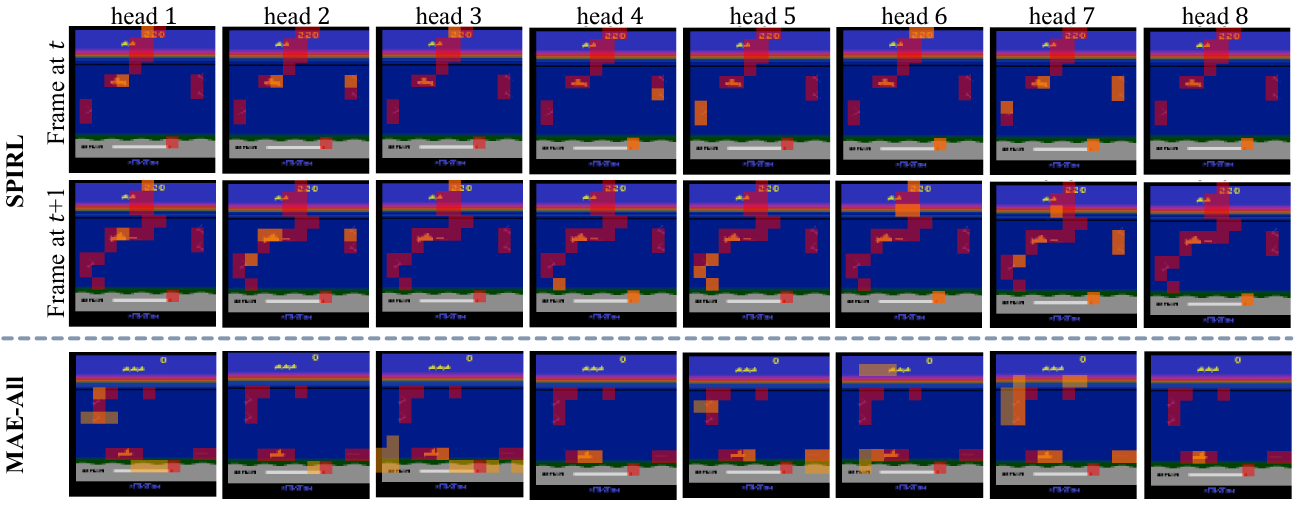}
    \ECMLzj{\vspace{-0.5cm}}
    \caption{\zj{Visualization for policy attention. Salient patches selected with dynamic-$\nK$ are highlighted with \pw{red squares}, while policy-attended patches are marked with transparent yellow color. Patches selected and attended are highlighted with \pw{orange squares}.}}
    \label{fig:interpretability_main}
    \ECMLzj{\vspace{-0.2cm}}
\end{figure}
}

\section{Conclusion}\label{sec:conclusion}

We presented a novel method to detect/select salient patches based on Vision Transformer without relying on convolutional features.
In contrast to keypoint-based methods, our technique is fast, requires fewer pre-training data, and allows adaptive and varying number of salient patches.
We showed how it can be exploited in RL to obtain a competitive method in the low-data training regime.

As future work, our salient patch selection technique could be applied on other downstream tasks.
Moreover, it would be interesting to address the discussed shortcomings of \ours{} (but also other keypoint-based or object-based methods), e.g., how to exploit the static background if it is important for decision-making? how to drop irrelevant moving background elements?
%
%

\TODO{\# pages: \thepage}


%
%
%
\bibliographystyle{splncs04}
\bibliography{main}





\newpage
\appendix
\onecolumn

\section{Pre-Training Details}\label{appendix:pretrain_details}
We adapt the implementation of the original MAE in two main aspects to make it compatible with salient patch selection and downstream RL tasks: architecture and datasets.

\subsection{Architecture and Training Schedule}\label{appendix:pretrain_archtecture}
The standard architectures designed in the original ViT and MAE papers \cite{dosovitskiy2020ViT,he2022MAE} have tens of millions of parameters, which are too huge to be applied directly in the RL domain efficiently (i.e., with respect to computational and memory costs). 
Accordingly, we use a much smaller encoder with about 162K parameters as shown in \cref{tab:compare_ViT_set}. 
Moreover, to enhance the ability for background reconstruction to select moving objects, we use a larger decoder than encoder, which is contrary to the original MAE design.

\begin{table}[h]
\caption{Comparison of architecture settings between \ours{} (for Atari game frames) and original MAE (for real image dataset like ImageNet) \cite{he2022MAE}. Note that to count the total number of parameters for MAE, we need to sum up the encoder and decoder's number of parameters, as well as the size of trainable [cls] token and [mask] token.}
\label{tab:compare_ViT_set}
\vskip 0.1in
\centering
\begin{small}
\begin{tabular}{ll|cccc}
    
        \toprule
        \multicolumn{2}{l}{\multirow{2}{*}{Hyper-parameter}} & \multirow{2}{*}{\ours{}} & \multicolumn{3}{c}{Original MAE} \\
        \multicolumn{2}{c}{} & & Base & Large & Huge \\
         \midrule
        \multicolumn{2}{l|}{Image shape} & \multicolumn{1}{c|}{(96, 96, 3)} & \multicolumn{3}{c}{(224, 224, 3)} \\
        \multicolumn{2}{l|}{Patch size} & \multicolumn{1}{c|}{8} & 16 & 16 & 14 \\
        \multicolumn{2}{l|}{\# Patches} & \multicolumn{1}{c|}{$12\times 12$} & $14\times 14$
        & $14\times 14$ & $16\times 16$ \\
        \midrule
        \multirow{3}{*}{Encoder} & Embed dim & \multicolumn{1}{c|}{64} & 768 & 1024 & 1280 \\
         & Depth & \multicolumn{1}{c|}{3} & 12 & 24 & 32 \\
         & \#Heads & \multicolumn{1}{c|}{4} & 12 & 16 & 16 \\
         & \#Parameters & \multicolumn{1}{c|}{162,432} & 85,646,592 & 303,098,880 & 630,434,560\\
         \midrule
        \multirow{3}{*}{Decoder} & Embed dim & \multicolumn{1}{c|}{128} &\multicolumn{3}{c}{512} \\
         & Depth & \multicolumn{1}{c|}{3} & \multicolumn{3}{c}{8} \\
         & \#Heads & \multicolumn{1}{c|}{8} & \multicolumn{3}{c}{16}\\
         & \#Parameters & \multicolumn{1}{c|}{628,160} & 26,007,808 & 26,138,880 & 26,177,612\\
        \midrule
        \multicolumn{2}{l|}{Total \#Parameters} & \multicolumn{1}{c|}{790,784} & 111,655,680 & 329,239,296 & 656,613,964\\
        \bottomrule
    \end{tabular}
\end{small}
\vskip -0.1in
\end{table}

We keep the same training schedule as the original MAE in general (\cref{tab:pretrain_MAE_details}), except that we use less training epochs and smaller number of batch size since our training datasets is much simpler and smaller than real image datasets like ImageNet \cite{deng2009imagenet}. 
\begin{table}[h]
\caption{Hyper-parameters for MAE pre-training on Atari games.}
\label{tab:pretrain_MAE_details}
\vskip 0.1in
\centering
\begin{small}
    \begin{tabular}{lcc}
    \toprule
    Hyper-parameter    & \multicolumn{2}{c}{Value}\\
    \midrule
    Mask ratio         & \multicolumn{2}{c}{75\%}  \\
    Batch size         & \multicolumn{2}{c}{64}    \\
    Epochs             & \multicolumn{2}{c}{50}   \\
    Warm up epochs        & \multicolumn{2}{c}{5}     \\
    Weight decay       & \multicolumn{2}{c}{0.05}  \\
    Base learning rate & \multicolumn{2}{c}{0.001} \\
    Learning rate      & \multicolumn{2}{c}{base\_learning\_rate * batch\_size / 256} \\
        Optimizer          & \multicolumn{2}{c}{AdamW($\beta_1=$0.9, $\beta_2=$0.95)} \\
    \bottomrule                         
    \end{tabular}
\end{small}
\vskip -0.1in
\end{table}

\subsection{Compare Pre-training Requirements with Baselines}\label{appendix:compare_pretraining_baselines}
RGB frames are collected from a random uniform policy that taking actions from a uniform distribution with the same environment wrapper as RL training (see \cref{tab:hyper_env_wrapper}). Pixel values are pre-processed by normalizing from [0,255] to [0,1]. \cref{tab:pretrain_num_frames} shows the number of frames to pre-train each game. Dataset will be randomly shuffled before training. 
\cref{tab:pretrain_num_frames} illustrates the length of trajectories for per-training. 
\begin{table}[ht]
\caption{Total number of frames for pre-training.}
\label{tab:pretrain_num_frames}
\vskip 0.1in
\centering
\begin{small}
    \begin{tabular}{lc}
    \toprule
    Game       & \# Frames\\
    \midrule
    Frostbite  & 5K \\
    MsPacman   & 50K \\
    Seaquest   & 50K \\
    BattleZone & 50K \\
    \bottomrule
    \end{tabular}
\end{small}
\vskip -0.1in
\end{table}

As demonstrated in \cref{tab:compare_pretrain}, our method has the least requirement for pre-training in terms of data quality and quantity\footnote{ About the length of trajectories needed for Transporter pre-training, the authors only said that 100K pairs of frames are needed, but their did not give the exact value of $len(\tau)$ to generate them. According to their description that a diverse dataset is needed and the re-implementation code from PermaKey, we think the value should be larger than \ours{}.} compare with other baselines. We could expect better performance if datasets with better quality (e.g., more frames from better rollouts) are used or online trained with RL.

\begin{table}[h]  
\caption{Comparison of the training settings between \ours{} and other methods. 
$len(\tau)$ denotes the length of trajectories for pre-training. The 3$^{rd}$ row compares use which kind of policy to collect trajectories, where \textit{Rand} refers to uniform random policy and \textit{Trained} refers to pre-trained policies with existing RL algorithms from Atari Model Zoo \cite{such2019atarizoo}. \zj{Moreover, since Transporter's pre-training data needs diversity, we mark it as \textit{Rand-Diverse}.} Fine-tune set to T means online training with RL policy is required and F means not.}
\label{tab:compare_pretrain}
\vskip 0.1in
\centering
\begin{small}
    \centering
    \begin{tabular}{llcccc}
        \toprule
        \multicolumn{2}{l|}{Config} & \ours & PermaKey & Transporter & MOREL \\
        \midrule
        \multirow{2}{*}{$\tau$} & \multicolumn{1}{|l|}{$len(\tau)$} & $<$50K & 85K & —— & 100K \\
        & \multicolumn{1}{|l|}{Policy} & \textit{Rand} & \textit{Trained} & \textit{Rand-Diverse} & \textit{Rand} \\
        \multicolumn{2}{l|}{Fine-tune} & F & T & F & T \\
        \bottomrule
    \end{tabular}
\end{small}
\vskip -0.1in
\end{table}

\subsection{Configuration to Compare \zj{Pre-training Performance}}\label{appendix:machine_config}
To visualize and compare the performance of salient part selection between \ours{} and baselines, we pre-train Transoprter and PermaKey from scratch with the same environment and machine configuration. Pre-training experiments are run on a single GPU card from a server with 4$\times$GeForce RTX 2070 Super 4$\times$7.8Gi GPU and 4$\times$64 GB memory. The configuration to produce \cref{fig:compare_kpts} is listed in \cref{tab:pretrain_config_time_cost}. We keep the same setting as in the published code from PermaKey\footnote{The authors have code for both PermaKey and Transporter} except that we use 16 as the pre-training batch size for PermaKey since it is memory costly and larger number of batch size can not be satisfied on our test machine.

\begin{table}[h]
\caption{Pre-training configuration and time cost.}
\label{tab:pretrain_config_time_cost}
\vskip 0.1in
\centering
\begin{small}
    \begin{tabular}{lccc}
    \toprule
          & Transporter & PermaKey & \ours{}\\
    \midrule
    $len(\tau)$  &  85K & 85K & $\le$50K \\
    Training epochs & 100 & 50 & 50 \\
    Batch size & 64 & 16 & 64\\
    Time(hours) & 25.9 & 50.7 & 1.6\\
    \bottomrule
    \end{tabular}
\end{small}
\vskip -0.1in
\end{table}

\zj{
\subsection{Visualization to Compare Salient-patches / Keypoints Selection}\label{appendix:compare_kps}
See \cref{fig:more_figs_kp}.
\begin{figure}[ht]
    \centering
    \includegraphics[width=\columnwidth]{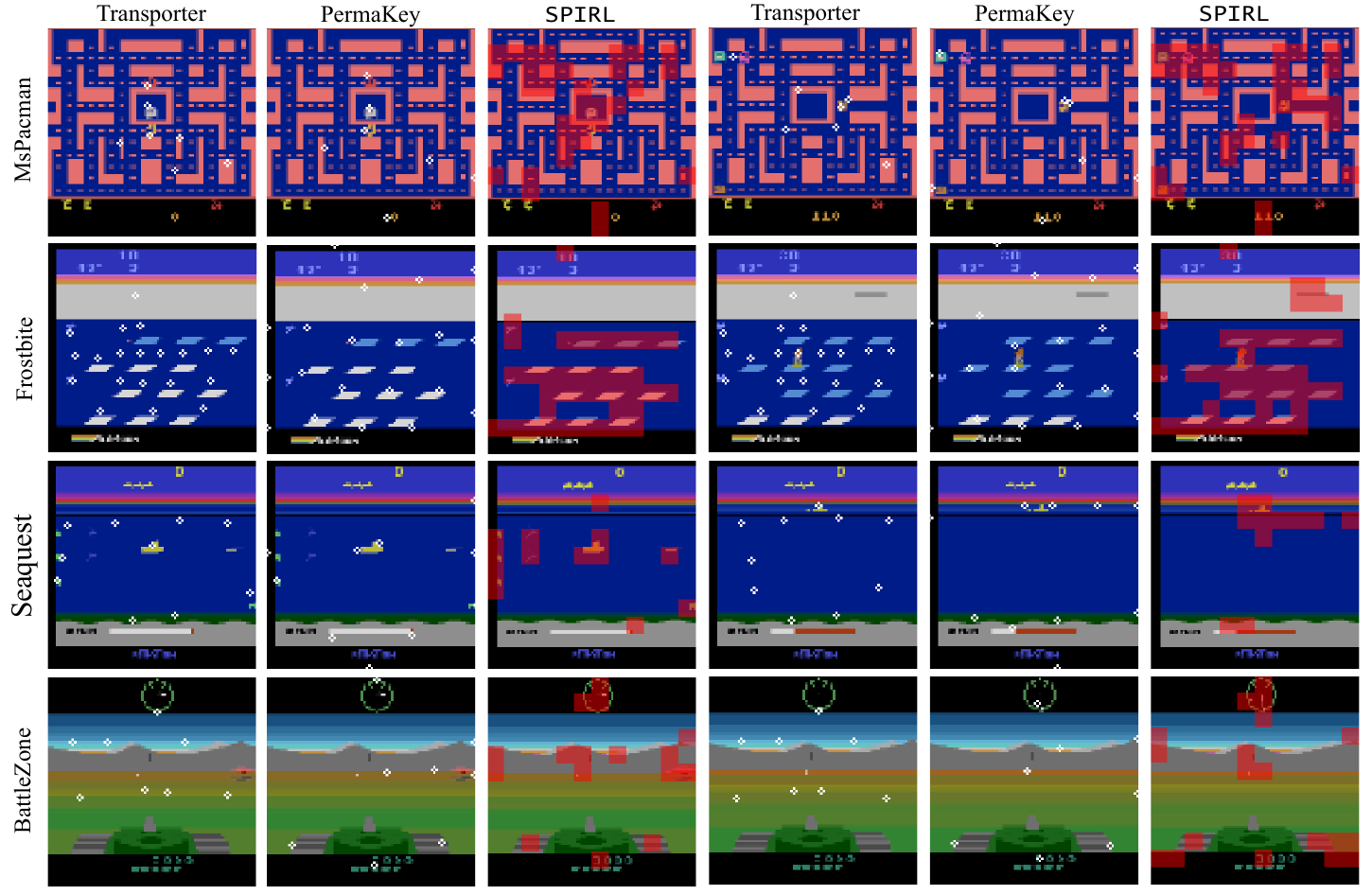}
    \caption{More visualization to compare \ours{} salient patch selection with baselines. The left 3 columns and right 3 columns corresponding to 2 different frames.}
    \label{fig:more_figs_kp}
\end{figure}
\TODO{more figures to visualize dynamic-K}
}
\section{RL Training Details}\label{appendix:rltrain_details}

\subsection{Environment Wrapper}\label{appendix:rl_training_env}
Our environment configurations follow \zj{standard requirements \cite{machado2018revisitALE}}, which is also adopted in baselines we compared with, except that we use RGB frames downsampled from shape $210\times160\times3$ to $96\times96\times3$. \cref{tab:hyper_env_wrapper} lists the related details.

\begin{table}[h]
\caption{Hyper-parameters for environment wrapper.}
\label{tab:hyper_env_wrapper}
\vskip 0.1in
\centering
\begin{small}
    \begin{tabular}{lc}
    \toprule
    Hyper-parameter           & Value       \\
    \midrule
    Grey-scaling              & False       \\
    Observation down-sampling & (96, 96)    \\
    Frame stacked             & 4           \\
    Frame skipped             & 4           \\
    Action repetitions         & 4           \\
    Max start no ops          & 30          \\
    Reward clipping           & {[}-1, 1{]} \\
    Terminal on loss of life  & True        \\
    Max frames per episode    & 108K        \\
    \bottomrule     
    \end{tabular}

\end{small}
\vskip -0.1in
\end{table}

\subsection{Rainbow Configurations}\label{appendix:rainbow_config}
We base our code on the original Rainbow implementation, but with modified hyperparameters introduced by DE-Rainbow as shown in \cref{tab:hyper_ourRainbow}. Compared with original Rainbow with CNN features of the whole frame as the input to its policy network, we use selected patch embeddings from the frozen pre-trained MAE encoder. We do not fine-tune the pre-trained encoder, therefore we can save the selected patch embeddings to replay buffer directly to reduce computational cost for off-policy RL training.

\begin{table}[h]
\caption{Hyper-parameters that differs from the original Rainbow setting \cite{hessel2018rainbow} used in our implementation.}
\label{tab:hyper_ourRainbow}
\vskip 0.1in
\centering
\begin{small}
    \begin{tabular}{lcc}
    \toprule
    Hyper-parameter & \multicolumn{2}{c}{For both \emph{100K} and \emph{400K} experiments}  \\
    \midrule
    Minimal steps to replay\ECMLzj{/update} & \multicolumn{2}{c}{1600} \\
    \ECMLzj{N-step} & \multicolumn{2}{c}{20} \\
    DQN hidden size & \multicolumn{2}{c}{256} \\
    Learning rate & \multicolumn{2}{c}{0.0001} \\
    $\#$Training updates & \multicolumn{2}{c}{100K} \\
    \midrule
    Hyper-parameter & \emph{100K} & \emph{400K}\\
    \midrule
    Priority $\beta$ increase steps & 100K & 400K \\
    Buffer size & 100K & 400K \\
    Steps per training update & 1 & 4 \\
    \bottomrule     
    \end{tabular}
\end{small}
\vskip -0.1in
\end{table}

\subsection{Evaluation Details}\label{appendix:evaluation_details}
\ECMLzj{For \cref{tab:exp_results_atari} and \cref{tab:compare_mmr_mr}, numbers without (resp. with) parentheses are average values (resp. standard deviation). Best results for each game are marked in bold.}
For each trial, average score and median score is tested with 50 different environments using different seeds. The experimental results are averaged over 5 trials. 
\subsection{Source of Experimental Results for \cref{tab:exp_results_atari}}\label{appendix:baselines_source}
The results of SimPLe, Transporter, and PermaKey come from their original papers. 
For PermaKey, we take the best results among their CNN and GNN versions. 

Since the original DE-Rainbow's \zj{\cite{van2019DERainbow}} code is not published, \zj{its} results are evaluated with our re-implementation by replacing \ours{}'s feature extractor with the data-efficient CNN as introduced in
\zj{their paper.}
Note that \ours{} already uses 
\zj{hyper-parameters introduced in \cite{van2019DERainbow}}
to configure the other RL part as explained in \cref{appendix:rainbow_config}. \cref{tab:compare_rainbow} compares the scores between our re-implemented Rainbow and the reported scores.
To make the evaluation more reliable, we use 10 trials while the original paper only uses 5.

\begin{table}[h!]
\caption{Comparison of our evaluation (averaged over 10 seeds) of DE-Rainbow against the performance reported in \zj{\cite{van2019DERainbow}}. 
}
\label{tab:compare_rainbow}
\vskip 0.1in
\centering
\begin{small}
    \begin{tabular}{lcc}
    \toprule
    Game & Reported & Ours  \\
    \midrule
    Frostbite & \textbf{866.8} & 341.4(277.8)\\
    MsPacman & \textbf{1204.1} & 1015.2(124.3)\\
    Seaquest & 354.1 & \textbf{396.5(124.4)}\\
    BattleZone & 10124.6 & \textbf{10602.2(2299.6)}\\
    \bottomrule     
    \end{tabular}
\end{small}
\vskip -0.1in
\end{table}

\zj{
As for DE-Rainbow-\ECMLzj{P}, \cref{tab:compare_derp} compares the different hyper-parameter settings:
(1) we set the minimal steps to replay for each game as the size of our pre-training dataset, which is a much larger value than the default (50K or 5K v.s. 1.6K);
(2) we keep the total number of learning steps and replay buffer size the same with default.
Since policy will use a total random policy to interact with environments before reaching minimal steps to replay, DE-Rainbow-\ECMLzj{P} equals to fill the replay buffer with the same data as our pre-training datasets.

\begin{table}[]
\caption{\zj{Compare hyper-parameter settings for DE-Rainbow-\ECMLzj{P} against DE-Rainbow and \ours{}.}}
\label{tab:compare_derp}
\begin{adjustbox}{max width=\textwidth}
    \begin{tabular}{lccc}
    \toprule
    \multirow{2}{*}{Configuration} & \multicolumn{2}{c}{DE-Rainbow-P}           & DE-Rainbow/\ours{} \\
    \cline{2-3}
                                  & MsPacman/BattleZone/Seaquest\hspace{0.1cm} & Frostbite & All Games                 \\
    \midrule
    Minimal steps to replay        & 50K                            & 5K        & 1.6K                 \\
    \ECMLzj{$\#$Env steps}         & 148.4K                         & 103.4K    & 100K                 \\
    Buffer size                    & 100K & 100K & 100K\\
    \bottomrule
    \end{tabular}
\end{adjustbox}
\end{table}
}


\subsection{Attention RL Implementation}\label{appendix:attention_RL_details}
\subsubsection{Architecture}
\zj{As shown in \cref{tab:attention_RL_details}}, our attention module includes a single transformer layer with pre-layer normalization, without residual connection inside the transformer layer, and use a trainable $[cls]$ as the pooling method. We kept the same positional embedding setting as MAE, which are pre-defined 2-D non-trainable sinusoid embeddings.
\begin{table}[h]
\caption{\zj{Configurations about our Transformer-based model architecture.}}
\label{tab:attention_RL_details}
    \centering
    \begin{tabular}{llc}
    \toprule
    \multicolumn{2}{l}{Configuration}            & Value                                 \\
    \midrule
    \multicolumn{2}{l}{Attention Depth}          & 1                                     \\
    \multicolumn{2}{l}{Number of Attention Head} & 8                                     \\
    \multicolumn{2}{l}{Projected Embedding Dim}  & 32                                    \\
    \multicolumn{2}{l}{Add Residual inside Attention Block}  & False                                    \\
    \multicolumn{2}{l}{Pooling Method}           & Trainable {$[cls]$} token                       \\
    \multicolumn{2}{l}{Positional Embedding}     & 2-D non-trainable sinusoid embeddings \\
    \midrule
    Configuration & Game & Value\\
    \midrule
    \multirow{4}{*}{Maximal Ratio}  & Frostbite  & 35\%                                  \\
                                    & MsPacman   & 30\%                                  \\
                                    & Seaquest   & 20\%                                  \\
                                    & BattleZone & 30\%  \\
    \bottomrule
    \end{tabular}
\end{table}


\subsubsection{Guidance to Choose the Maximal Ratio $\MR$}\label{appendix:select_mmr}
\begin{figure}[h]
    \centering
    \includegraphics[width=\textwidth]{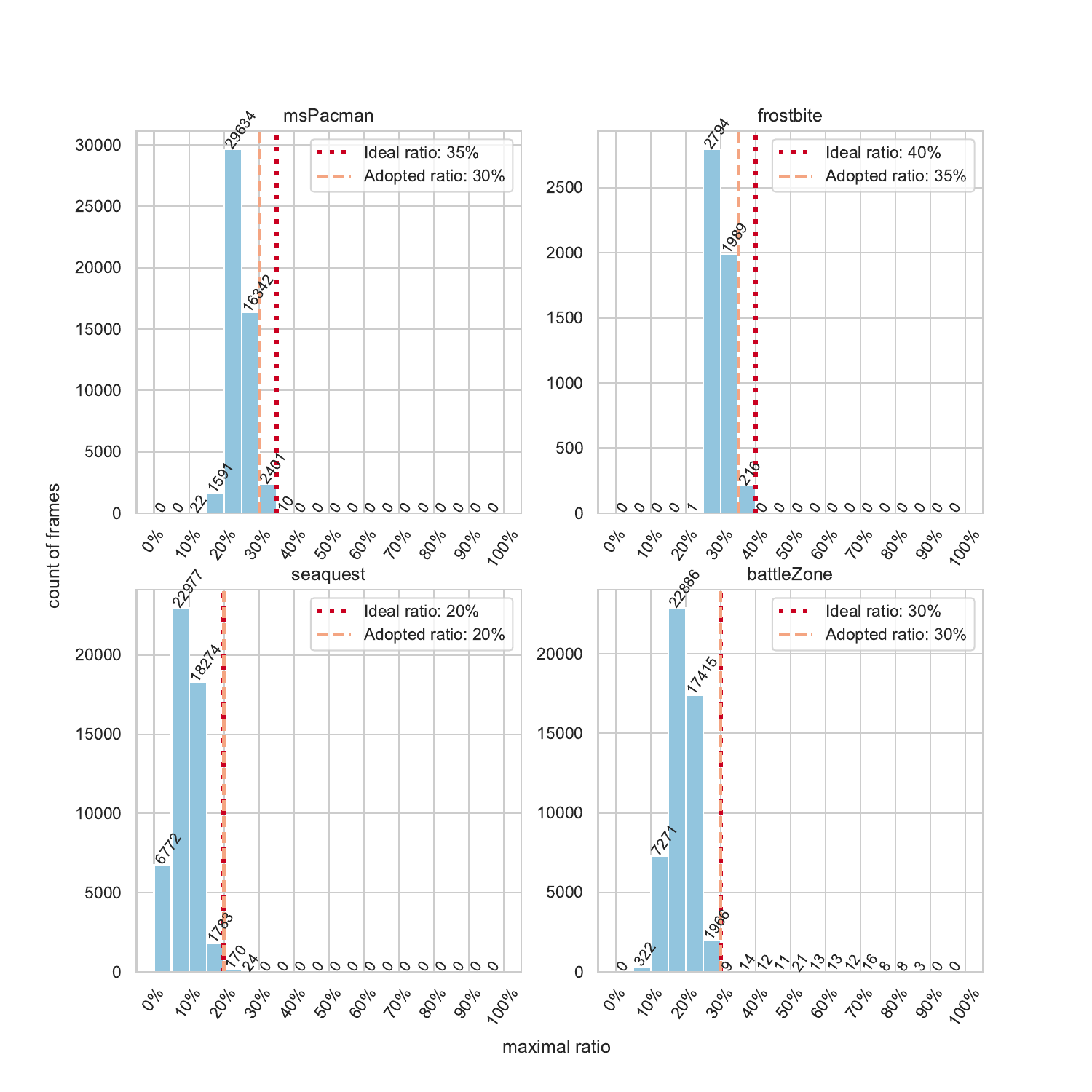}
    \caption{\zj{Histograms based on the count of salient patches in frames from the pre-training datasets of 4 games.}}
    \label{fig:select_mmr}
\end{figure}

\zj{
We exploit the pre-training dataset to provide a guidance to determine a suitable $\MR$ by counting the number of selected salient-patches of frames in datasets. Since we don't want to lose information about salient information, we choose an ideal maximal ratio from a discrete set $\{5\%, 10\%, \dots, 90\%, 95\%\}$ as $\idealMR$ that can guarantee salient maintenance for more than $99.9\%$ of frames in pre-training datasets. Practically we try $\{\idealMR-5\%, \idealMR, \idealMR+5\%\}$ and select the best one as the adopted $\MR$.
As shown in \cref{fig:select_mmr}, the ideal ratio $\idealMR$ is selected by guaranteeing more than $99.9\%$ will not lose salient information, while the adopted ratio is selected from $\{\idealMR-5\%, \idealMR, \idealMR+5\%\}$ according to the actual performance.
}

\section{More Experimental Results} \label{appendix:expe}

\subsection{Different Solutions to Deal with Dynamic Number of Embeddings for Transformer-based RL}\label{appendix:solution_attentionRL}
\zj{
We mainly test 3 different solutions on \emph{Seaquest}:
(1) Zero-Padding, which add dummy patches with zero padding when the number of selected patches is less than the maximal number;
(2) Trainable-$[pad]$, which use a trainable $[pad]$ token to replace non-salient-patches' embedding before inputing RL part;
(3) Masked-Attention, which use zeros to replace the softmaxed attention score for non-salient-patches inside the policy's Transformer blocks. 
\cref{tab:solutions_dynamic_RL} shows the efficiency of Zero-Padding solution, which is also the easiest to implement one.

\begin{table}[h]
    \centering
    \caption{\zj{Comparison of different solutions to deal with varying number of embedding inputs for RL. Tested on \emph{Seaquest} under \emph{100K} setting.}}
    \label{tab:solutions_dynamic_RL}
    \begin{adjustbox}{max width=\textwidth}
        \begin{tabular}{lccc}
        \toprule
        Method & Zero-Padding & Trainable-$[pad]$ & Masked-Attention \\
        \midrule
        Score  & \textbf{557.9(148.1)} & 491.6(139.6)              & 494.40(154.8)  \\
        \bottomrule
        \end{tabular}
    \end{adjustbox}
\end{table}
}

\subsection{Ablation study about Transformer-based RL Architecture}\label{appendix:abla_rl_arch}
\zj{

Instead of using a trainable class token $[cls]$ in the Transformer-based aggregation in the RL part, we also try a simple average pooling. 
\cref{tab:abla_attention} shows the ablation study about the pooling methods. 
Interestingly, the average pooling performs very well in the \emph{100K} setting, but $[cls]$ pooling reveals its strength in the \emph{400K} setting.
}
\begin{table}[h]
\caption{Aggregation methods in Transformer-based RL. \zj{We use a unified maximal ratio (50\%) for fair ablation.}}
\label{tab:abla_attention}
    \centering
    \begin{tabular}{lcccc}
        \toprule
        \multirow{2}{*}{Games} & \multicolumn{2}{c}{\emph{100K}, average} & \multicolumn{2}{c}{\emph{400K}, median} \\ \cmidrule(lr){2-3} \cmidrule(lr){4-5}
         & average & $[cls]$ & average & $[cls]$  \\ 
        \midrule
        Frostbite              & \textbf{604.3(788.7)}  & 425.7(563.3) &  689.0(950.8) & \textbf{1407.1(1312.4)}\\
        MsPacman               & 916.3(124.1)  & \textbf{957.6(236.2)} & 1113.0(187.5) & \textbf{1176.0(209.3)}\\
        Seaquest               & 507.8(82.4) & \textbf{541.1(89.1)}    & 550.0(178.6) & \textbf{552.0(97.8)}  \\
        BattleZone             & \textbf{10412.0(2535.1)} & 8447.1(1491.3) & 12900.0(2012.4) & \textbf{13500.0(1870.8)}\\
        \bottomrule
        \end{tabular}
\end{table}


\end{document}